\def\year{2022}\relax
\documentclass[letterpaper]{article} 
\usepackage{aaai22}  
\usepackage{times}  
\usepackage{helvet}  
\usepackage{courier}  
\usepackage[hyphens]{url}  
\usepackage{graphicx} 
\usepackage{natbib}  
\usepackage{caption} 
\DeclareCaptionStyle{ruled}{labelfont=normalfont,labelsep=colon,strut=off} 
\usepackage{algorithm}
\usepackage[noend]{algorithmic}

\usepackage{newfloat}
\usepackage{listings}
\floatstyle{ruled}
\newfloat{listing}{tb}{lst}{}
\floatname{listing}{Listing}
\title{Machine Learning for Utility Prediction in\\Argument-Based Computational Persuasion}

\author {
    Ivan Donadello\textsuperscript{\rm 1} 
    Anthony Hunter\textsuperscript{\rm 2}
    Stefano Teso\textsuperscript{\rm 4}
    Mauro Dragoni\textsuperscript{\rm 3} \\
}
\affiliations {
    \textsuperscript{\rm 1} Free University of Bozen-Bolzano, Italy \\
    \textsuperscript{\rm 2} University College London, United Kingdom \\
    \textsuperscript{\rm 3} Fondazione Bruno Kessler, Italy \\
    \textsuperscript{\rm 4} University of Trento, Italy \\
    ivan.donadello@unibz.it, anthony.hunter@ucl.ac.uk, stefano.teso@unitn.it, dragoni@fbk.eu
}

\usepackage{amssymb}
\usepackage{environ}
\usepackage{todonotes}
\usepackage{amsthm}
\usepackage{amsmath}
\usepackage{bbm}
\usepackage{bm}
\usepackage{booktabs}
\usepackage{multicol}
\usepackage{multirow}
\usepackage{tikz}
\usetikzlibrary{arrows, fit, shapes, automata}

\DeclareMathOperator*{\argmin}{argmin}
\newtheorem{definition}{Definition}
\newcommand{\children}{{\sf Children}}
\newcommand{\isdecision}{\mathrm{isDecision}}

\newcommand{\isnotleaf}{\mathrm{isNotLeaf}}
\newcommand{\rt}{\mathrm{root}}
\newcommand{\append}{\mathrm{append}}
\newcommand{\height}{{\sf height}}
\newcommand{\leaves}{{\sf leaves}}
\newcommand{\nodes}{{\sf Nodes}}

\newcommand{\amax}{{\sf AMax}}
\newcommand{\reals}{\mathbb{R}}

\usepackage{xcolor}

\begin{document}

\maketitle
\begin{abstract}
Automated persuasion systems (APS) aim to persuade a user to believe something by entering into a dialogue in which arguments and counterarguments are exchanged. To maximize the probability that an APS is successful in persuading a user, it can identify a global policy that will allow it to select the best arguments it presents at each stage of the dialogue whatever arguments the user presents. However, in real applications, such as for healthcare, it is unlikely the utility of the outcome of the dialogue will be the same, or the exact opposite, for the APS and user. In order to deal with this situation, games in extended form have been harnessed for argumentation in Bi-party Decision Theory. This opens new problems that we address in this paper: (1) How can we use Machine Learning (ML) methods to predict utility functions for different subpopulations of users? and (2) How can we identify for a new user the best utility function from amongst those that we have learned? To this extent, we develop two ML methods, EAI and EDS, that leverage information coming from the users to predict their utilities. EAI is restricted to a fixed amount of information, whereas EDS can choose the information that best detects the subpopulations of a user. We evaluate EAI and EDS in a simulation setting and in a realistic case study concerning healthy eating habits. Results are promising in both cases, but EDS is more effective at predicting useful utility functions.
\end{abstract}

\section{Introduction}
\label{sec:intro}
Persuasion is an activity that involves one party trying to induce another party to believe something. It is an important and multifaceted human facility. Often, it can involve a dialogue in which arguments and counterarguments are exchanged between the persuader and the persuadee. For example, a doctor can use arguments to persuade a patient to eat a more healthy diet, and the patient can give counterarguments based on their understanding and preferences. Also, the doctor may provide counterarguments to attempt to overturn misconceptions held by the patient.

An automated persuasion system (APS) is a software system that takes on the role of a persuader, and a user is the persuadee \cite{Hunter2019ki}. It aims to use convincing arguments in order to persuade the user. Whether an argument is convincing depends on its own nature, on the context of the argumentation, and on the user's characteristics. An APS maintains a model of the user to choose the best moves in the dialogue \cite{HadouxHunter2019}. During the turn of the APS, it presents an argument to the user and then user can select their counterarguments from a menu. An APS can also use a natural language interface to allow users input their argument in free text \cite{Chalaguine2020}.


In order for an APS to choose the moves it makes in the dialogue (i.e., the arguments it presents), it needs to use a strategy. Within the literature on computational models of arguments, key approaches to strategies are: 
{\bf Mechanism design} where it is a one-step process rather for multi-step dialogues \cite{RahwanLarson08,RL09,FanToni12};
{\bf Planning systems} to optimize choice of arguments based on belief in premises \cite{BCB14,BCH2017}
or minimize the number of moves made \cite{Atkinson2012}; 
{\bf Machine learning} of strategies such as  Reinforcement Learning \cite{Monteserin2012,Rosenfeld2016ecai,Rach2018,Katsumi2018,Riveret2019,Alahmari2020}
and transfer learning \cite{Rosenfeld2016}; 
{\bf Probabilistic methods} to select a move based on what an agent believes the other is aware of \cite{RienstraThimmOren13}, to approximately predict the argument an opponent might put forward based on data about the moves made by the opponent in previous dialogues \cite{HadjinikolisSiantosModgilBlackMcBurney13},
using a decision-theoretic lottery \cite{HunterThimm2016ijar},
using POMDPs when there is uncertainty about the internal state of the opponent \cite{HadouxBeynierMaudetWengHunter15};
and
{\bf Local strategies} based for example on the concerns (i.e., issues raised or addressed by an argument) of the persuadee \cite{HadouxHunter2019,Chalaguine2020}.

Apart from local strategies based on concerns, the above proposals do not take the user's needs into account. Yet, if an APS is to persuade someone to accept some arguments, then how those arguments relate to their needs is fundamental in deciding how to argue with them. This can be captured by considering the utility of each argument from the point of view of both the APS and the user (i.e., we have a utility function for each agent). For this we can harness games in extensive form \cite{Osborne1994}. For argumentation, this has led to Bi-party Decision Theory (BDT) \cite{HadouxHP18} that generates a policy that is a good compromise in maximizing the APS and user utilities. However, BDT requires a utility function to be constructed for groups of users. We address this by using Machine Learning (ML) methods that learn utility functions from data about subpopulations of users.
Then, an appropriate utility function for a new user (i.e., which subpopulation the user belongs to) has to be determined. This can be performed by asking questions. However, a user might be reluctant to answer too many questions (e.g., because they become bored, or it takes too much effort). So the challenge is to ask fewer questions and still get a useful utility function.
We address these novel problems by studying the following research questions: \textbf{RQ1} How do we use ML-methods to identify a utility function for each subpopulation of a set of users? \textbf{RQ2} How do we optimize the number of questions we ask a new user to identify an appropriate utility function?

Differently from \textbf{Recommender Systems} (RSs) \cite{2015rsh}, our proposal is based on BDT (Section \ref{sec:bdt}) and, therefore, it does not provide ``one-shot'' recommendations based only on users' utilities but a dialogue that considers both users' and system (i.e., biparty) utilities. BDT explicitly models the persuader's and user's needs. This allows a better customization of the APS according to the specific domain and subpopulation. {\bf Conversational RSs} (CRSs) \cite{JannachMCC21} go beyond standard ``one-shot'' RSs by creating a dialogue with users. However, as far as we know, no method tackles the problem of learning a utility function in a BDT setting for CRSs. {\bf Reinforcement Learning} (RL) \cite{Rosenfeld2016ecai} is orthogonal to our proposal but it is data inefficient as it involves large amounts of interaction. Our method is less data hungry as it only requires some utility elicitation from users with questionnaires. We addresses the limitations of BDT by developing two ML methods (Section \ref{sec:ML}) that compute the utility functions from a dataset of user needs. These ML methods allow the optimization of the number of questions to ask. The evaluation using synthetic data (Section \ref{sec:evaluation}) and a case study (Sections \ref{sec:case_study}) reveals promising results and answers to both RQ1 and RQ2.

\section{Bi-party Decision Theory}
\label{sec:bdt}
\begin{figure}
\begin{center}
\begin{tikzpicture}[-,thick]
\node[draw,rectangle,fill=yellow] (n1)  at (3.5,5) {\scriptsize $n_1$};
\node[draw,rectangle,dashed,fill=yellow] (n2)  at (3.5,3.5) {\scriptsize $n_2$};
\node[draw,rectangle,fill=yellow] (n3)  at (1.5,2.1) {\scriptsize $n_3$};
\node[draw,rectangle,fill=yellow] (n4)  at (6,2.1) {\scriptsize $n_4$};
\node[fill=yellow] (n5)  at (0,0) {\scriptsize $n_5$};
\node[fill=yellow] (n6)  at (3,0) {\scriptsize $n_6$};
\node[fill=yellow] (n7)  at (5,0) {\scriptsize $n_7$};
\node[fill=yellow] (n8)  at (7.5,0) {\scriptsize $n_8$};
\path (n2)[] edge[] node[draw,rectangle,rounded corners,line width=0.3pt,text width=45mm,font=\scriptsize,fill=gray!20]{[$a_1$] Low red meat consumption is necessary for a healthy diet.} (n1);
\path (n3)[] edge[] node[draw,rectangle,rounded corners,line width=0.3pt,left,text width=25mm,font=\scriptsize,fill=gray!20]{[$a_2$] It is really difficult to change diet.} (n2);
\path (n4)[] edge[] node[draw,rectangle,rounded corners,line width=0.3pt,right,text width=25mm,font=\scriptsize,fill=gray!20]{[$a_3$] I really like the taste of meat.} (n2);
\path (n5)[] edge[] node[draw,rectangle,rounded corners,line width=0.3pt,pos=0.5,text width=20mm,font=\scriptsize,fill=gray!20]{[$a_4$] Think about the benefits of reducing red meat.} (n3);
\path (n6)[] edge[] node[draw,rectangle,rounded corners,line width=0.3pt,pos=0.5,right,text width=12mm,font=\scriptsize,fill=gray!20]{[$a_5$] Try to reduce red meat slowly.} (n3);
\path (n7)[] edge[] node[draw,rectangle,rounded corners,line width=0.3pt,pos=0.5,left,text width=15mm,font=\scriptsize,fill=gray!20]{[$a_6$] White meat can be an alternative.} (n4);
\path (n8)[] edge[] node[draw,rectangle,rounded corners,line width=0.3pt,pos=0.5,text width=15mm,font=\scriptsize,fill=gray!20]{[$a_7$] Fish is a tasty alternative to meat.} (n4);
\end{tikzpicture}
\end{center}
\caption{A decision tree for a persuasive dialogue for the red meat persuasion goal. Proponent (decision) nodes are solid boxes and opponent (chance) nodes are dashed boxes.}
\label{fig:example}
\end{figure}
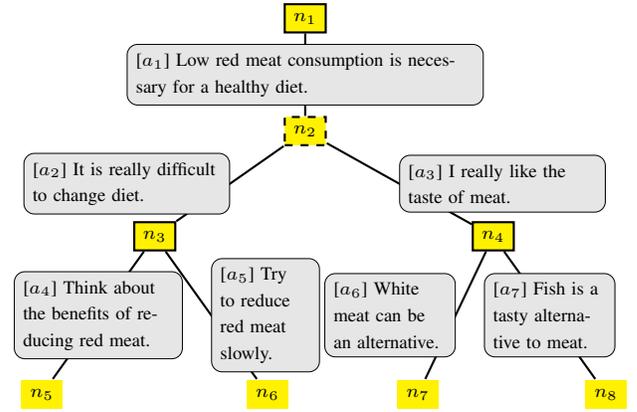

Decision trees (DT) are an important approach to making optimal decisions \cite{Peterson1994}. In computational persuasion, a DT is used for representing all possible dialogues between two agents: each path from the root to a leaf alternates \emph{decision nodes} with \emph{chance nodes}. The former are associated with the persuader (a.k.a. the proponent) and represent the argument to be posed by the APS. The latter are associated with the persuadee (a.k.a. the opponent) and represent the argument posed by the user. Each arc $(n,n_i)$ is labelled with the argument posited by the corresponding agent that has been selected at node $n$. Figure \ref{fig:example} contains an example of DT for the persuasion goal of reducing the red meat consumption.
A \emph{policy} associates a decision node with the best argument to pose by the proponent by considering the points of view of both the proponent and the opponent. In Bi-party Decision Theory, these viewpoints are modelled using two \emph{utility functions} $U^p$ and $U^o$. These provide each leaf with a utility value that represents the benefit, for the corresponding agent, of accepting the APS arguments in the dialogue from the root to that leaf 

A {\bf bimaximax policy} maximizes $U^p$ at a decision node and $U^o$ at a chance node. Some notation is necessary before a formal definition of the policy. Let $T$ be a DT, $L$ be a labeling function that associates an argument to each arc $(n, n_i)$ between two nodes of $T$ and $\children(T,n)$ be the set of children of $n$. The $\amax(T,U,n)$ function returns the children of $n$ with highest utility $U$:
$
\amax(T,U,n) = \{ n' \in {\children}(T,n) \mid \forall n'' \in {\children}(T,n) , U(n') \geq U(n'') \}
$
Let $\delta \in \mathbb{R}$ be a discount factor that decreases the utility of longer branches. Indeed, shorter dialogues can be more persuasive than longer ones that require more attention from the user. Definition \ref{def:bimaximax} defines the bimaximax policy.
\begin{definition}
\label{def:bimaximax}
A {\bf bimaximax policy} for $(T,L,U^p,U^o,\delta)$ is $\Pi: \nodes(T) \rightarrow  \nodes(T)$ defined as follows using the calculation of the $Q^p: \nodes(T)\rightarrow \reals$ and $Q^o: \nodes(T)\rightarrow \reals$ functions.
\begin{itemize}
\item If $n$ is a leaf node, then $Q^p(n) = U^p(n)$ and $Q^o(n) = U^o(n)$.
\item If $n$ is a chance node, and $n_i \in  \amax(T,Q^o,n)$, then $Q^o(n) = \delta \times Q^o(n_i)$ and $Q^p(n) = \delta \times Q^p(n_i)$.
\item If $n$ is a decision node, and $n_i \in  \amax(T,Q^p,n)$, then $Q^o(n) = \delta \times Q^o(n_i)$ and $Q^p(n) = \delta \times Q^p(n_i)$ and $\Pi(n) = \langle n_i, L(n,n_i) \rangle$.
\end{itemize}
\end{definition}
For the example in Figure \ref{fig:example}, the APS aims to persuade the user to adopt healthy alternatives to red meat, the leaf utility values are then $U^p(n_5) = 2$, $U^p(n_6) = 3$, $U^p(n_7) = 4$, $U^p(n_8) = 6$. The user is interested in consuming white meat and to slowly reduce his/her consumption of red meat: $U^o(n_5) = 2$, $U^o(n_6) = 6$, $U^o(n_7) = 7$, $U^o(n_8) = 4$. The utility values are then propagated to the higher levels of the tree as follows, where $Q^{p/o}(n)$ denotes $(Q^p(n),Q^o(n))$: $Q^{p/o}(n_1) = (3,6)$, $Q^{p/o}(n_2) = (3,6)$, $Q^{p/o}(n_3) = (3,6)$, $Q^{p/o}(n_4) = (6,4)$, with $\delta=1$, $\Pi(n_1) = (n_2, L(n_2))$, $\Pi(n_3) = (n_6, L(n_6))$ and $\Pi(n_4) = (n_8, L(n_8))$.

Definition \ref{def:bimaximax} does not provide a criterion for the choice of $n_i$ when  $|\amax(T,Q,n)| > 1$. We therefore adopt a random choice that improves the variability of the APS messages.

\subsection{Simulated Dialogues}
The bimaximax policy defines a rule for choosing the label/argument to pose given a decision node $n$. Given a chance node $n_i$, the next decision node $n_j$ is selected by the user via, for example, a menu. However, when real users are not available the next decision node has to be computed according to some \emph{simulated opponent policy} $\Pi^o$. This policy can be computed by statistical methods that rely on datasets of conversations with users, by harvesting arguments from the web or with the use of questionnaires. Here, we simply define the opponent policy as $\Pi^o(n_i) = \langle n_j, L(n_i, n_j)\rangle$ with $n_j \in  \amax(T,Q^o,n_i)$. In our example $\Pi^o(n_2) = (n_3, L(n_3))$. We implement Definition \ref{def:bimaximax} in a procedure, called $\mathrm{Bimaximax}$, that propagates up the leaf utilities at the parents, with a post-order tree traversal, and outputs $\Pi$, $\Pi^o$. A \emph{simulated dialogue procedure}, Algorithm \ref{algo:dialogue}, consists in the computation of the policies $\Pi, \Pi^o$ (line 2) and in alternating between the proponent and opponent arguments (given by $\Pi, \Pi^o$) from the root to a leaf (lines from 4 to 9). The output path $\bm{p}$ contains the proponent/opponent nodes from the root to a leaf. In our example, $\bm{p} = \langle n_1, n_2, n_3, n_6\rangle $.
\begin{algorithm}[h!tb]
\caption{$\mathrm{SimDialogue}(T,L,\bm{u}^p,\bm{u}^o,\delta)$}
\label{algo:dialogue}
\textbf{Input}: $T,L,\bm{u}^p,\bm{u}^o,\delta$\\
\textbf{Output}: The path $\bm{p} = \langle \rt(T), n_1, \ldots, n_{\height(T)}\rangle$\\
\begin{algorithmic}[1] 
\STATE $\bm{p} := \langle \rangle$
\STATE $\Pi, \Pi^o := \mathrm{Bimaximax}(T,L,\bm{u}^p,\bm{u}^o,\delta)$
\STATE $n := \rt(T)$
\WHILE{$\isnotleaf(n)$}
\IF {$\isdecision(n)$}
\STATE $\langle n, l_n \rangle := \Pi(n)$
\ELSE
\STATE $\langle n, l_n \rangle := \Pi^o(n)$
\ENDIF
\STATE $\bm{p}.\append(n)$ 
\ENDWHILE
\STATE \textbf{return} $\bm{p}.\append(n)$
\end{algorithmic}
\end{algorithm}

Let $\bm{u}^p = \langle U^p(n_1), U^p(n_2), \ldots \rangle$ and $\bm{u}^o = \langle U^o(n_1)$, $U^o(n_2), \ldots \rangle$ be the proponent and opponent utility vectors respectively, with $n_i \in \leaves(T)$. The values of $\bm{u}^p$ can be authored by domain experts from the substantial evidence in healthcare literature on the health benefits of specific behaviour changes, e.g., the value of changing from high red meat consumption to white meat consumption  \cite{abete2014association}. Instead, the setting of $\bm{u}^o$ is an open challenge. In the next section, we define two ML methods able to predict an accurate estimation for $\bm{u}^o$ from a dataset of users' utilities. These methods can be used also in a dialogue procedure with real users where the policy $\Pi^o$ will be replaced, for example, by a menu in a smartphone app that implements the APS.

\section{Machine Learning for Utility Prediction}
\label{sec:ML}
Let $T$ be a DT, a corresponding utility dataset $\mathcal{U}^o$ contains the utility values for a population sample. We encode $\mathcal{U}^o$ as a matrix where each row $j$ contains the utility vector $\bm{u}^o_j$ of the arguments in $\leaves(T)$ for a given user. Such a dataset can be gathered by using, for example, utility elicitation techniques based on questionnaires \cite{lenert1997measurement, lenert2001utility}.
Here, we assume that $\mathcal{U}^o$ contains samples of subpopulations (as commonly done in RSs), that is, clusters of users who have similar utilities. For example, according to the leaf arguments in Figure \ref{fig:example}, we can have three subpopulations/clusters in $\mathcal{U}^o$: $C_1$, $C_2$ and $C_3$. The users in $C_1$ like to eat food containing animal proteins, the users in $C_2$ have difficulties changing their diet habits whereas the users in $C_3$ are more open minded. Such a dataset can be gathered with cluster sampling techniques \cite{turk2005review}.

We develop two ML methods, trained on $\mathcal{U}^o$, for predicting an estimation of the opponent utility vector $\hat{\bm{u}}^o$, for a given user, as close as possible to the true vector $\bm{u}^o$. This implies choosing some features of the users as input to our ML methods to predict $\hat{\bm{u}}^o$. These can be demographic information or some arguments in $\leaves(T)$. We consider these input utilities as the \emph{evidence} to be obtained in order to predict the rest of the utility vector. Therefore, the predicted utility vector is composed from the evidence $\bm{u}_{1:e}^o$, with $e \in [1, Le -1]$ and $Le = |\leaves(T)|$, and the predicted utilities: $\hat{\bm{u}}_e^o = \langle \bm{u}_{1:e}^o;\hat{\bm{u}}_{e+1:Le}^o \rangle$. In our example, let us suppose that a user in cluster $C_1$ has true utilities $\bm{u}^o = \langle 5,4,8,9 \rangle$. A ML method, trained on $\mathcal{U}^o$, can ask as evidence the utilities for $n_5, n_6$ ($e=2$ out of 4 possible questions) to predict the utilities of $n_7, n_8$: 9 and 7, for example. Therefore, $\hat{\bm{u}}_2^o = \langle 5,4,9,7 \rangle$ and this is a good prediction as the true path returned by $\mathrm{SimDialogue}$ is $\bm{p} = \langle n_1, n_2, n_4, n_8\rangle$ whereas the path returned by using $\hat{\bm{u}}_2^o$ is $\hat{\bm{p}} = \langle n_1, n_2, n_4, n_7\rangle$ that is quite similar to the true one. Indeed, both $n_7$ and $n_8$ contain arguments related to the animal proteins cluster.

Our proposal consists of providing $\mathrm{SimDialogue}$ with the opponent utility vectors $\hat{\bm{u}}_e^o$ predicted with ML methods as input. The evidence $e$ obtained from the user has to be enough to have a good dialogue path $\hat{\bm{p}}$ but without asking too many questions of the user (otherwise they may disengage). For example, asking 50 utilities of a DT with 60 leaves will bring reliable paths but would be too many questions. More formally, let $\mathrm{ML}$ be a machine learning method trained on $\mathcal{U}^o$ to compute the opponent utility vector $\hat{\bm{u}}_e^o$ from $\bm{u}^o_{1:e}$, that is, $\mathrm{ML}(\bm{u}^o_{1:e}) = \hat{\bm{u}}_e^o$, and let $\mathcal{R}$ be a regret function that measures the difference from the path returned by $\mathrm{SimDialogue}$ with the one returned by $\mathrm{SimDialogue(ML)}$, here the notation $\mathrm{SimDialogue(ML)}$ is a shortcut for $\mathrm{SimDialogue}(T,L,\bm{u}^p,\mathrm{ML}(\bm{u}^o_{j,1:e}), \delta))$, more details in Section \ref{sec:metrics}.  Given a test set $\mathcal{U}_{TE}^o$, the optimal number $e^*$ of evidence to ask about is:
\begin{multline}
\label{eq:opt}
e^* = \argmin_{e \in [1, Le-1]}\sum_{j=1}^{|\mathcal{U}_{TE}^o|} \mathcal{R}(\mathrm{SimDialogue}(T,L,\bm{u}^p,\bm{u}_j^o, \delta),\\ \mathrm{SimDialogue}(T,L,\bm{u}^p,\mathrm{ML}(\bm{u}^o_{j,1:e}), \delta)) + \mathcal{E}(e)
\end{multline}
with $\mathcal{E}$ a monotonic-increasing function that models the user effort in answering the asked evidence whose parameters can be estimated, for example, by measuring the user disengagement rate. This is to avoid having an APS that asks all the possible questions as evidence. The following subsections present our ML methods for utility prediction.

\subsection{Evidence as Amount of Information} 
An \emph{evidence as amount of information} (EAI) method uses a portion of the arguments in $\mathcal{U}^o$ (limited by the evidence $e$) to train a supervised regressor to predict the utility values of the other arguments. This is the standard \emph{regression} task performed with \emph{supervised learning} techniques where a training set $\mathcal{X}\times \mathcal{Y}$ is used to compute an estimator function $f$ for $\mathcal{Y}$. The set $\mathcal{X}$ contains feature vectors $\bm{x}$ whereas $\mathcal{Y}$ contains numeric values $y$ to be estimated from $\bm{x}$. The training is performed such that the output $\hat{y}$ of $f(\bm{x})$ has to be as close as possible to the true value $y$. In our case, given an evidence index $e$, the set $\mathcal{X}$ is given by the first $e$ arguments in $\mathcal{U}^o$ and $\mathcal{Y}$ is one of the remaining arguments. More formally, the training set is $\mathcal{U}^o_{*,1:e} \times \mathcal{U}^o_{*,e+l}$ with $l \in [1, Le -e]$. The notation $\bm{A}_{*, i:j}$ indicates the columns from $i$ to $j$ (included) of matrix $\bm{A}$. Therefore, for each evidence index $e \in [1, Le -1]$ multiple regressions have to be performed from $\mathcal{U}^o_{*,1:e}$: one for predicting the utility values for each remaining argument with index in $[e+1, Le]$. As the ML method for computing $f$, we use a \textbf{Support Vector Regression} $\mathrm{SVR}$ that linearly penalizes the predictions $\hat{y}$ that are at least $\epsilon$ away from the true value $y$. The advantage of an EAI method is that many off-the-shelf ML methods for computing $f$ can be used. However, such methods are restricted to the provided portion $\mathcal{U}^o_{*,1:e}$ of the dataset to learn the other utility values, without any possibility of automatically choosing a better set of arguments to ask about. The next method addresses this issue. We opted for $\mathrm{SVR}$ (and $\mathrm{RF}$ in the next section) because it is a known-working technique that does not require large training set to reach reasonable performance. Our focus is not on the choice of the ML method but on the usefulness of adopting ML for effective, data driven, argumentation.

\subsection{Evidence as Depth of Searching}
An \emph{evidence as depth of searching} (EDS) method has a limited amount of evidence $e$ to be obtained from users, but, differently from the EAI-based approach, the EDS method is given access to the whole dataset $\mathcal{U}^o$. Therefore, the EDS method can search the most appropriate evidence for the regression of the utility function. This search is limited by the quantity $e$. We call the method \textbf{Cluster and RAndom forest MEan Regressor} ($\mathrm{CRAMER}$) and, in brief, it performs clustering on $\mathcal{U}^o$ to find subpopulations. Then a Random Forest (RF) is trained to learn the most important utilities of arguments to ask in order to classify a new user into a subpopulation. The utility vector $\hat{\bm{u}}_e^o$ is given by the asked utilities and by the centroids of the subpopulation. We detail the method with the running example.

Firstly, $\mathrm{CRAMER}$ performs clustering on $\mathcal{U}^o$ and discovers the underlying clusters and computes their centroids. These are the mean of the utilities of each user in the cluster. Numeric examples for the centroids for $C_1, C_2$ and $C_3$ are: $\bm{c}(C_1) = \langle 5.0,4.0,9.2,9.1 \rangle$, $\bm{c}(C_2) = \langle 3.1,8.5,7.2,1.9 \rangle$ and $\bm{c}(C_3) = \langle 8.1,8.6,2.7,7.3 \rangle$.
The second step of $\mathrm{CRAMER}$ is to train a RF classifier on $\mathcal{U}^o$ to learn from user utilities what are the main arguments that characterize a cluster.
RF classifiers require a max depth parameter for setting the maximum depth of their classification trees, i.e., decision trees trained for classification tasks. We set this with $e$, that is the number of arguments to be asked. A high value for max depth could generate classification trees that overfit $\mathcal{U}^o$ (with consequently low performance) and could be too demanding for the users. In our example, if we set $e=2$, a random forest is able to classify a new user by simply asking the utilities only for $n_8$ (fish) and $n_7$ (white meat). Indeed, high utilities for both these arguments mean that the user is in $C_1$, whereas a low utility for the fish argument classifies the user in $C_2$. A high utility for the fish and a low one for the white meat argument bring the classification to $C_3$. Lastly, the method joins the asked (true) utilities with the centroid (inferred) utilities in the predicted cluster to obtain the utility vector $\hat{\bm{u}}_e^o$ of the user. For example, the utility vector of a user answering with utility value of 8 for the white meat argument and with utility 7 for the fish argument (cluster $C_1$) is $\hat{\bm{u}}_2^o = \langle 5, 4, 8, 7\rangle$.

\section{Empirical Evaluation}
\label{sec:evaluation}
The aim of the evaluation is to test the ability of $\mathrm{SimDialogue(ML)}$ (prediction of $\hat{\bm{u}}^o$ with $\mathrm{ML}$ and consequent use of $\hat{\bm{u}}^o$ in Algorithm \ref{algo:dialogue}) of returning persuasive arguments as good as the ones returned by $\mathrm{SimDialogue}$. However, as big datasets of APS dialogues with real users are not available, we test $\mathrm{SimDialogue(ML)}$ on synthetic datasets with the use of simulations. We use the output of the $\mathrm{SimDialogue}$ procedure as a \emph{gold standard} and evaluate how much the output of $\mathrm{SimDialogue(ML)}$ differs from the gold standard.
The source code and the supplementary material are online at \url{https://github.com/ivanDonadello/ML-Argument-Based-Computational-Persuasion}.

\subsection{Simulation Design}
We compare $\mathrm{SimDialogue}$ and $\mathrm{SimDialogue(ML)}$ on different abstract DTs and population samples with their own utilities. Our aim is to perform a fair comparison between the gold standard and $\mathrm{SimDialogue(ML)}$ trying to avoid possible bias. Good simulation results reveal possible good results in real cases too. The simulation steps are as follows:

{\textbf{Trees Generation:}} given a tree height and a list of branching factors as input, a random abstract DT is generated with a breadth-first algorithm. This recursively generates random children of a node according to a branching factor taken from the input list. This is repeated until the input height has been reached. We repeat this process to obtain a set $\mathbb{T} = \{T_1, T_2, \ldots \}$ of abstract DTs. As they have no particular meaning, the labelling $L$ is not necessary. Each DT represents the structure of a possible persuasion dialogue between an APS and a user.

{\textbf{Datasets Generation:}} for each $T \in \mathbb{T}$, we synthesize a set $\mathbb{U}_T^o = \{\mathcal{U}_{T,1}^o, \mathcal{U}_{T,2}^o, \ldots\}$ of datasets of opponent utilities. Each dataset $\mathcal{U}_{T,i}^o = \{\bm{u}^o_{T,i}\}_j$ represents samples of subpopulations containing the utility vector $\bm{u}^o_{T,i,j}$ for a user indexed by $j$ belonging to a particular subpopulation. This allows the EAI and EDS methods to learn utility functions for subpopulations. Each dataset is used for: i) training by procedure $\mathrm{SimDialogue(ML)}$, ii) testing by both $\mathrm{SimDialogue(ML)}$ and $\mathrm{SimDialogue}$. Several datasets of sobpopulations allow us to check the results of $\mathrm{SimDialogue(ML)}$ in a robust way. If we tested our method on only one dataset, the results could be affected by the quality of that particular dataset.
    
We synthesize a dataset $\mathcal{U}_{T,i}^o$ by: i) creating clusters of users; ii) assigning each user a utility vector $\bm{u}_{T,i,j}^o$ such that users in the same cluster have similar utilities. 
Let $T$ be a DT, $C \in \mathbb{N}$ be an input number of clusters and $CW \in \mathbb{R}$ be an input center width, i) we sample $C$ center points (vectors with $\leaves(T)$ dimensions) from a multivariate uniform distribution with values in $[-CW,CW]$. These vectors are the centers of the clusters. ii) For each center vector $\bm{c}$, we sample $m$ vectors (with $\leaves(T)$ dimensions) from a multivariate normal distribution with mean $\bm{\mu} = \bm{c}$ and covariance matrix $\bm{\Sigma} = \sigma_C^2\bm{I}$, with $\bm{I}$ the identity matrix and $\sigma_C^2$ the cluster variance given as input. We sample the $m$ vectors such as every cluster has roughly the same number of vectors. All these sampled vectors compose the dataset $\mathcal{U}_{T,i}^o$.

{\textbf{Simulation Run:}} for each pair $\langle T, \mathcal{U}_{T,i}^o \rangle$ and for a given evidence index $e \in [1, Le - 1]$, we split $\mathcal{U}_{T,i}^o $ into a training set, for training a given $\mathrm{ML}$ method, and a test set. Let $j$ be the index of a sample in the test set, we run $\mathrm{SimDialogue}(T,L,\bm{u}^p, \mathrm{ML}(\bm{u}_{T,i,j,1:e}^o),\delta)$ and $\mathrm{SimDialogue}(T,L,\bm{u}^p,\bm{u}_{T,i,j}^o,\delta)$. Remember that an EAI method takes a portion of the training set given by $e$, whereas an EDS method takes the whole training set but can ask only $e$ questions.
For a statistical significance of the results, we use the $k$-fold cross validation technique. The dataset $\mathcal{U}_{T,i}^o$ is split into $k$ parts, $k-1$ parts are used as training set for $\mathrm{SimDialogue(ML)}$ and the remaining part is left as test set for both $\mathrm{SimDialogue(ML)}$ and $\mathrm{SimDialogue}$. In this way, $k$ splits/folds of the original dataset $\mathcal{U}_{T,i}^o$ are obtained and for each split we run both $\mathrm{SimDialogue(ML)}$ and  $\mathrm{SimDialogue}$.

{\textbf{Comparison:}} the paths $\hat{\bm{p}}_{T,i,j,e,k}$ and $\bm{p}_{T,i,j,k}$ returned by the $\mathrm{SimDialogue(ML)}$ and $\mathrm{SimDialogue}$ procedures, respectively, are compared according to the following metrics.

\subsection{Metrics for the Empirical Evaluation}
\label{sec:metrics}
Given a user in $\mathcal{U}^o$, let $\hat{\bm{u}}_e^o$ be their estimated utility vector and $\bm{u}^o$ be the true one. Given $\hat{\bm{p}} = \mathrm{SimDialogue}(T,L,\bm{u}^p,\hat{\bm{u}}_e^o,\delta)$ and $\bm{p} = \mathrm{SimDialogue}(T,L,\bm{u}^p,\bm{u}^o,\delta)$ (hereafter we remove the subscripts $T, i,e,k$ if not necessary), we consider the last (leaf) nodes of these paths: $\hat{n} = \hat{\bm{p}}_h$, $n = \bm{p}_h$ with $h$ the height of $T$, we call $n$ the \emph{true node} and $\hat{n}$ the \emph{predicted node} and we compute the evaluation metrics by only comparing $n$ with $\hat{n}$. This strategy is motivated by the random choice of $n_i$ when  $|\amax(T,Q,n)| > 1$: this randomness can easily select a different branch in $T$ for $\mathrm{SimDialogue(ML)}$ w.r.t. $\mathrm{SimDialogue}$ for the same utility vector. This side effect would highly penalize the performance if we compared the whole paths resulting in a non-fair comparison. 

The aim of the evaluation is to test the ability of $\mathrm{SimDialogue(ML)}$ to return similar nodes to $\mathrm{SimDialogue}$. This similarity can be easily tested with the standard accuracy, i.e., whether $n_j=\hat{n}_j$ for a given user indexed with $j$ in $\mathcal{U}^o$. However, as random choices can easily fail this testing, the similarity between $n_j$ and $\hat{n}_j$ is relaxed by measuring an \emph{argument distance} between the arguments in $n_j$ and in $\hat{n}_j$, respectively. This is the $\mathcal{R}$ term in Equation \eqref{eq:opt}. The nodes $n_j$ and $\hat{n}_j$ have a small argument distance if the opponent and proponent utilities of the argument in $n_j$ are close to the corresponding ones of the argument in $\hat{n}_j$. Good performance would indicate that an APS based on $\mathrm{SimDialogue(ML)}$ is able to provide arguments as similar as possible (w.r.t. the utilities) to the gold standard arguments returned by $\mathrm{SimDialogue}$. More formally, the \textbf{mean argument distance} ($Mad$) is the mean, over all the dataset samples, of the Manhattan distance between the proponent and opponent utilities of the true node and the predicted one:
$Mad = \frac{1}{|\mathcal{U}^o|}\sum_{j=1}^{|\mathcal{U}^o|}|U^p(n_j) - U^p(\hat{n}_j)| + |U^o(n_j) - U^o(\hat{n}_j)|.$
Notice that, if we decompose the mean argument distance according to the proponent ($\frac{1}{|\mathcal{U}^o|}\sum_{j=1}^{|\mathcal{U}^o|}|U^o(n_j) - U^o(\hat{n}_j)|$) and the opponent ($\frac{1}{|\mathcal{U}^o|}\sum_{j=1}^{|\mathcal{U}^o|}|U^p(n_j) - U^p(\hat{n}_j)|$) dimensions we obtain the \emph{mean absolute errors} (Mae) which are standard metrics for regression tasks. These metrics depend on $T, i, e, k$, therefore the output of the simulations is a set of 
argument distances. These values need to be aggregated to have a global estimation of the performance.

{\bf $\mathbf{k}$-fold aggregation} The first aggregation is performed on the number of folds $k$ of the $k$-fold cross validation. For each triple $\langle T, i, e \rangle$ we compute the mean and the standard deviation of the metrics over all the $k$ values obtaining the values $\mu(Mad)_{T,i,e}$ and $\sigma(Mad)_{T,i,e}$.

{\bf Evidence aggregation} The second aggregation regards the evidence $e$ requested as input to users in order to compute the utility values of the remaining arguments. We want to study the trend of the performance metrics with a increasing amount of evidence. This amount is measured as the percentage of asked utility arguments over all possible arguments/leaves of a DT. This allows us to check whether there exists a percentage of evidence from which the performance is acceptable. We therefore aggregate the mentioned means by averaging over $\mathbb{T}$ and $\mathbb{U}_T^o$ according to the evidence percentage. Let $Ep \in \{0.1, 0.2, \ldots 1.0\}$ be the evidence percentage and $E = \left \lceil{Ep\cdot(Le-1)})\right \rceil$ be the amount of utility of arguments to ask for $T$ and for a given $Ep$, the mean argument distance aggregation is the average of the set $\{\mu(Mad)\}_{T,i,e}$, with $e \in \{1, 2, \ldots E\}.$ The standard deviation of this aggregations is computed from this set by estimating the pooled variance.


\subsection{Simulation Setting}
\label{sec:setting}
Some values of the parameters in the simulation are uniformly (randomly) selected from a set of alternatives. This avoids the bias of having only one value resulting in more general results. These are: the tree height in $\{4,5,6\}$, the proponent utility in $\{1,2, \ldots 11\}$, the number of clusters $C \in \{4, 6, 8, 10\}$ and the center width $CW \in \{0.5, 1.0, 2.5, 3.0\}$. This setting for $CW$, allow us to have different levels of clustering performance avoiding the bias of population samples that are perfectly separable into clusters. The branching factor for generating the children of a node is 2 or 3 for 90\% of the time, 4 for 10\% of the time. This allows us to have realistic DTs with a reasonable number of leaves. Other parameters have a single value: the number of synthetic trees ($|\mathbb{T}|=10$) and datasets ($|\mathbb{U}^o_T|=10$), the size of $\mathcal{U}^o_{T,i}$ is 2000, the cluster variance $\sigma^2_C$ is 1.0, the discount factor $\delta$ in $\mathrm{Bimaximax}$ is 1 as it is not relevant for the simulations and $k=5$ for the $k$-fold cross validation. The hyperparameters for $\mathrm{SVR}$ are $C=1$, $\epsilon = 0.1$ and the radial basis function as a kernel. For the clustering in $\mathrm{CRAMER}$ we used KMeans with the number of clusters found through grid search in $\{4,6,8,10\}$. The random forest in $\mathrm{CRAMER}$ has 100 estimators with the minimum number of samples required: i) to split a node is 2, ii) to be a leaf is 1.

\subsection{Results}
\begin{figure*}[h!tbp]
\centering
\includegraphics[width=0.84\textwidth]{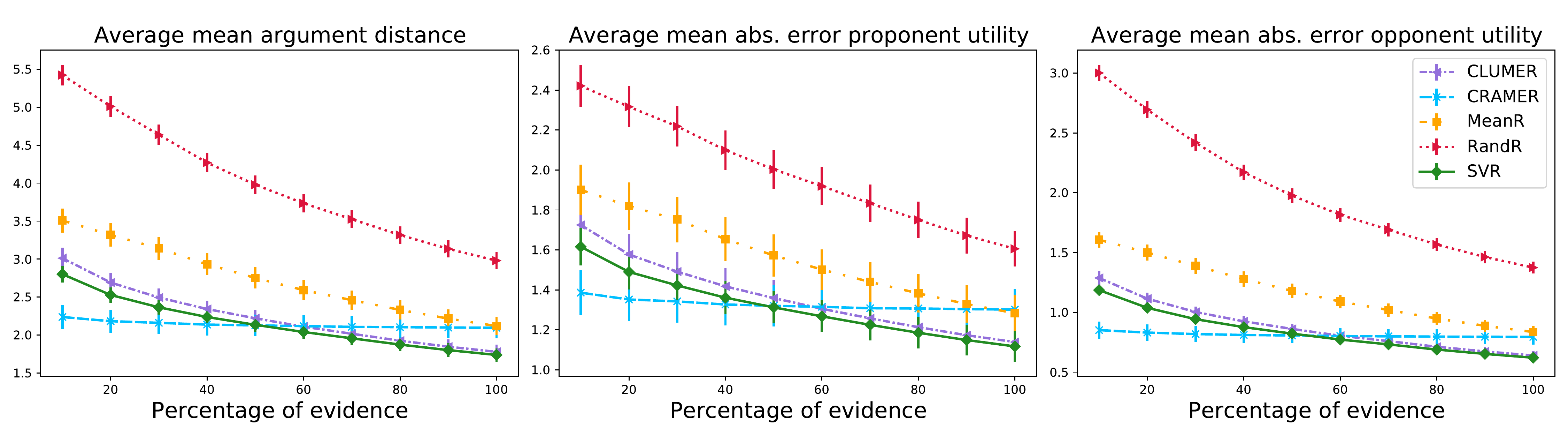}
\caption{Performance of $\mathrm{SimDialogue(ML)}$. Results are aggregated according to the trees, the datasets and the evidence. Vertical bars represent the sampled standard deviation. Best viewed in colors.}
\label{fig:sim_results}
\end{figure*}
We measure the performance of $\mathrm{SimDialogue(ML)}$ (with the average mean argument distance) according to different levels of evidence percentage $E_p$. This allows us to check: i) whether ML methods are able to learn utility functions from data in comparison to non-ML baselines (RQ1) and ii) whether there exists a percentage of evidence from which the performance are acceptable (RQ2). We compare some baselines with the gold standard in the same way done for $\mathrm{SimDialogue(ML)}$:
\begin{itemize}
\item \textbf{Random Regressor} ($\mathrm{RandR}$) randomly selects integer utility values between the maximum and the minimum utility for each argument in the columns of of $\mathcal{U}^o_{*,e+1:Le}$.
\item \textbf{Mean Regressor} ($\mathrm{MeanR}$) computes the mathematical mean (over all the users) of the utility values of each argument in the columns of $\mathcal{U}^o_{*,e+1:Le}$.
\item \textbf{CLUster and MEan Regressor} ($\mathrm{CLUMER}$) performs clustering to find the subpopulations in $\mathcal{U}^o_{*,1:e}$. Once a new user is assigned to a cluster, the centroids of the clusters are used as utility values for the remaining arguments in the columns of $\mathcal{U}^o_{*,e+1:Le}$.

\end{itemize}
Aggregated performance are reported in Figure \ref{fig:sim_results} whereas Appendix A of the supplementary material provides examples of performance for some random DTs and datasets. As expected, the EAI method and the baselines increase their performance as the evidence percentage $E_p$ increases. Indeed, more asked questions allow more accurate performance. In addition, the increase of the performance is also due to a robustness property of the bimaximax policy: i.e., the system ability to capture regions of high user's utility regardless the other regions. If few leaves have a random utility most probably their utility will not be the highest one. Indeed, during the bottom-up utility propagation phase in $\mathrm{Bimaximax}$, the utility of these few leaves will be discarded in favour of higher utilities from other branches as the bimaximax policy selects the highest utility at each node. These few leaves do not have the power to change the prediction of the true node. As $E_p$ increases, the number of random utilities decreases. This explains also the unexpected performance improvement of $\mathrm{RandR}$ as $E_p$ increases.

{\bf Discussion for RQ1} 
The results for the argument distances are promising as with 10\% of evidence $\mathrm{SVR}$ obtains a $Mad$ of 2.8. This means that the predicted node has proponent and opponent utilities that on average have distance 1.4 from the true node. As the utility values range in a Likert scale from 1 to 11, a 1.4 distance is an acceptable error. Therefore, the ML methods are able to compute the utility values for the population sample by predicting nodes that contain arguments that are very close (from the utilities point of view) to the arguments contained in the true nodes.

{\bf Discussion for RQ2} An acceptable number of questions as evidence is not trivial to identify. For example, 10\% of evidence could present low performance for small trees (e.g., with 20 leaves only 2 questions are asked) therefore choosing 30\% could represent a better choice. However, this could result in a high amount of requested information (e.g., 30 questions for a tree with 100). To this extent, $\mathrm{CRAMER}$ gives us a different view. For this method, 10\% of evidence is required to identify the subpopulation and the utilities are the centroids of each subpopulation. Therefore, for a tree with 20 leaves, a depth of 2 for the random forest is sufficient to identify 4 subpopulations. In general, a depth of $D$ identifies $2^D$ subpopulations. It is important to notice that the level of depth of search is a \emph{maximum} level for the random forest, that is, the random forest can use less information to identify subpopulations. This can be seen in the picture as the performance of $\mathrm{CRAMER}$ are stable as the depth of search increases. The EAI methods need 50\% of evidence to have the same performance of $\mathrm{CRAMER}$.

The advantage of an EDS method relies in its ability of automatically selecting the most appropriate arguments to ask from $\mathcal{U}^o$ obtaining good performance with a low $E_p$. Differently, an EAI method has only a partial view of the whole information but it increases the performance with higher $E_p$. The drawback of $\mathrm{CRAMER}$ is its rough computation of $\hat{\bm{u}_e^o}$ given a subpopulation. $\mathrm{SVR}$ instead has a more accurate computation based on the error optimization of the predicted utility. This can be seen by its better performance with respect to the EAI method $\mathrm{CLUMER}$. In addition, $\mathrm{CLUMER}$ and $\mathrm{SVR}$ have similar performance as the former explicitly leverages the cluster structure of the users, the latter achieves the same through the underlying Gaussian kernel. 

\section{A Realistic Case Study}
\label{sec:case_study}
\begin{figure*}[h!tbp]
\centering
\includegraphics[width=0.84\textwidth]{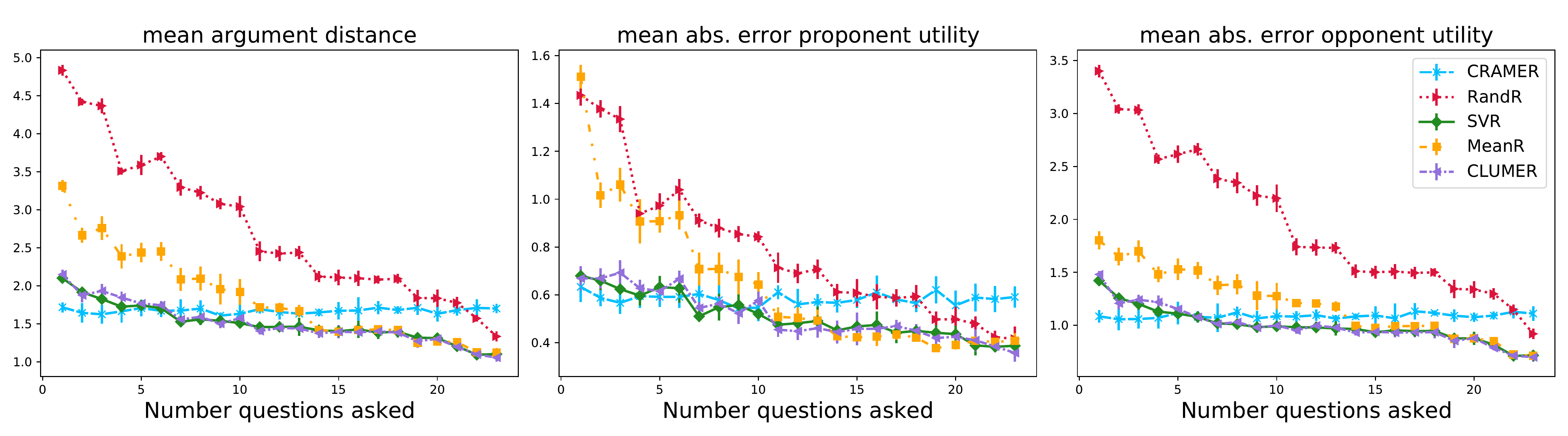}
\caption{Performance of $\mathrm{SimDialogue(ML)}$ for the red meat case study. Results are aggregated according to the $k$ folds. Vertical bars represent the standard deviation. Best viewed in colors.}
\label{fig:meat_results}
\end{figure*}
We evaluate $\mathrm{SimDialogue(ML)}$ on a realistic case study where the DT contains real arguments (taken from the literature) regarding the reduction of red meat consumption. Both the DT and the synthetic dataset are provided. This is the first step for implementing an APS with a smartphone app for following healthy lifestyles. This was commissioned by the Local Healthcare Department of Trento (Italy), for the deployment in the Trentino area.

Regarding the DT, we gathered the arguments and their relations from diet and behavior change literature \cite{abidi2014d,stankevitz2017perceived}. We obtained a DT with 23 leaves and height of 4. Concerning $\bm{u}^p$, we labeled each leaf with a topic describing the content of the arguments in the corresponding path from the root. In our example, the path $n_1, n_2, n_4, n_8$ is labeled with ``Fish as alternative''. Each topic is assigned with a proponent utility according to the topic importance from a diet point of view. Each leaf $n$ is assigned with the proponent utility of the corresponding topic. These topics are: vegetarianism ($U^p(n) = 8$), fish as alternative ($U^p(n) = 6$), white meat as alternative ($U^p(n) = 4$), thinking of alternatives ($U^p(n) = 3$) and reducing red meat consumption ($U^p(n) = 2$).

Regarding the dataset, we created 6 user profiles with their demographics (sex, age, school degree, level of meat consumption and level of physical activity) and utility values according to the profile, e.g, young student or woman in career.
Each profile is then transformed into a numeric vector.
This dataset of 6 examples is expanded to 2000 examples by repeating and shuffling the 6 example vectors. Then a matrix of gaussian random noise ($\mu =0$, $\sigma^2=1$) is added to the expanded dataset in order to have variability. DT topics and dataset profiles are in Appendix C.


The evaluation has been performed as in the previous section. As there is only one DT and dataset, we present the non-averaged metrics in Figure \ref{fig:meat_results}. This realistic case study shows similar findings to the simulations. However, differently from the simulations, here demographic information is available as evidence for all the methods. 
The $Mad$ is good as with only the demographic information we have a $Mad$ of around 1.7 for $\mathrm{CRAMER}$ and around 2.1 for both $\mathrm{SVR}$ and $\mathrm{CLUMER}$. Here, the simple clustering and centroids approach of $\mathrm{CLUMER}$ is sufficient to achieve the same results of $\mathrm{SVR}$. $\mathrm{CRAMER}$ presents stable performance. As the number of questions increases both $\mathrm{CLUMER}$ and $\mathrm{SVR}$ require 5 questions (out of 22 possible questions, i.e., 23\% more of questions) to get a $Mad$ close to the one of $\mathrm{CRAMER}$. The mean absolute error of the proponent utility is much lower compared to the opponent one. This is due to a bias in the dataset as most of the final true nodes have 8 as proponent utility value. This bias does not regard the opponent utility, see the statistic in Appendix D. A qualitative comparison of the arguments returned by  $\mathrm{CLUMER}$ and $\mathrm{SVR}$ for each profile is in Appendix E.

\section{Conclusion}
\label{sec:conclusion}
In computational persuasion, Bi-party Decision Theory is a promising approach for an APS to choose the best arguments to persuade a user. These strategies are based on utility functions. However, no methods deal with the construction of such functions for a new user in an efficient way. In this paper, we addressed this problem by developing two ML models (EAI and EDS) to learn these utility functions from datasets. The evaluation with simulations and with a realistic case study show that both EAI and EDS are able to learn utility functions of subpopulation of users with comparable performance. However, EDS is more efficient as it requires less user information.

As future work, we want to test these methods with real users in a living lab. In addition, a combination of EAI with EDS could improve the performance requiring a lower number of questions to ask about. The use of neural networks for learning the utilities will be studied also in an interactive APS that it adapts the utilities as it interacts with the user \cite{DragoneTP18}.

\section* {Acknowledgements}
The research of ST was partially supported by TAILOR, a project funded by EU Horizon 2020 research and innovation programme under GA No 952215.

\bibliography{main}

\begin{thebibliography}{32}
\providecommand{\natexlab}[1]{#1}

\bibitem[{Abete et~al.(2014)Abete, Romaguera, Vieira, de~Munain, and
  Norat}]{abete2014association}
Abete, I.; Romaguera, D.; Vieira, A.~R.; de~Munain, A.~L.; and Norat, T. 2014.
\newblock Association between total, processed, red and white meat consumption
  and all-cause, CVD and IHD mortality: a meta-analysis of cohort studies.
\newblock \emph{British Journal of Nutrition}, 112(5): 762--775.

\bibitem[{Abidi et~al.(2014)Abidi, Vallis, Abidi, Piccinini-Vallis, and
  Imran}]{abidi2014d}
Abidi, S.; Vallis, M.; Abidi, S. S.~R.; Piccinini-Vallis, H.; and Imran, S.~A.
  2014.
\newblock D-WISE: Diabetes Web-Centric Information and Support Environment:
  conceptual specification and proposed evaluation.
\newblock \emph{Canadian Journal of Diabetes}, 38(3): 205--211.

\bibitem[{Alahmari(2020)}]{Alahmari2020}
Alahmari, S. 2020.
\newblock \emph{Reinforcement Learning for Argumentation}.
\newblock Ph.D. thesis, University of York, York, UK.

\bibitem[{Atkinson, Bench{-}Capon, and Bench{-}Capon(2012)}]{Atkinson2012}
Atkinson, K.; Bench{-}Capon, P.; and Bench{-}Capon, T. J.~M. 2012.
\newblock Efficiency in Persuasion Dialogues.
\newblock In \emph{{ICAART} {(2)}}, 23--32. SciTePress.

\bibitem[{Black, Coles, and Bernardini(2014)}]{BCB14}
Black, E.; Coles, A.~J.; and Bernardini, S. 2014.
\newblock Automated Planning of Simple Persuasion Dialogues.
\newblock In \emph{{CLIMA}}, volume 8624 of \emph{Lecture Notes in Computer
  Science}, 87--104. Springer.

\bibitem[{Black, Coles, and Hampson(2017)}]{BCH2017}
Black, E.; Coles, A.~J.; and Hampson, C. 2017.
\newblock Planning for Persuasion.
\newblock In \emph{{AAMAS}}, 933--942. {ACM}.

\bibitem[{Chalaguine and Hunter(2020)}]{Chalaguine2020}
Chalaguine, L.; and Hunter, A. 2020.
\newblock A Persuasive Chatbot Using a Crowd-Sourced Argument Graph and
  Concerns.
\newblock In \emph{{COMMA}}, volume 326 of \emph{FAIA}, 9--20. {IOS} Press.

\bibitem[{Dragone, Teso, and Passerini(2017)}]{DragoneTP18}
Dragone, P.; Teso, S.; and Passerini, A. 2017.
\newblock Constructive Preference Elicitation.
\newblock \emph{Frontiers Robotics {AI}}, 4: 71.

\bibitem[{Fan and Toni(2012)}]{FanToni12}
Fan, X.; and Toni, F. 2012.
\newblock Mechanism Design for Argumentation-based Persuasion.
\newblock In \emph{{COMMA}}, volume 245 of \emph{Frontiers in Artificial
  Intelligence and Applications}, 322--333. {IOS} Press.

\bibitem[{Hadjinikolis et~al.(2013)Hadjinikolis, Siantos, Modgil, Black, and
  McBurney}]{HadjinikolisSiantosModgilBlackMcBurney13}
Hadjinikolis, C.; Siantos, Y.; Modgil, S.; Black, E.; and McBurney, P. 2013.
\newblock Opponent Modelling in Persuasion Dialogues.
\newblock In \emph{{IJCAI}}, 164--170. {IJCAI/AAAI}.

\bibitem[{Hadoux et~al.(2015)Hadoux, Beynier, Maudet, Weng, and
  Hunter}]{HadouxBeynierMaudetWengHunter15}
Hadoux, E.; Beynier, A.; Maudet, N.; Weng, P.; and Hunter, A. 2015.
\newblock Optimization of Probabilistic Argumentation with Markov Decision
  Models.
\newblock In \emph{{IJCAI}}, 2004--2010. {AAAI} Press.

\bibitem[{Hadoux and Hunter(2019)}]{HadouxHunter2019}
Hadoux, E.; and Hunter, A. 2019.
\newblock Comfort or Safety? Gathering and Using the Concerns of a Participant
  for Better Persuasion.
\newblock \emph{Argument \& Computation}, 10: 113--147.

\bibitem[{Hadoux, Hunter, and Polberg(2018)}]{HadouxHP18}
Hadoux, E.; Hunter, A.; and Polberg, S. 2018.
\newblock Biparty Decision Theory for Dialogical Argumentation.
\newblock In \emph{{COMMA}}, volume 305 of \emph{Frontiers in Artificial
  Intelligence and Applications}, 233--240. {IOS} Press.

\bibitem[{Hunter et~al.(2019)Hunter, Chalaguine, Czernuszenko, Hadoux, and
  Polberg}]{Hunter2019ki}
Hunter, A.; Chalaguine, L.; Czernuszenko, T.; Hadoux, E.; and Polberg, S. 2019.
\newblock Towards Computational Persuasion via Natural Language Argumentation
  Dialogues.
\newblock In \emph{{KI}}, volume 11793 of \emph{LNCS}, 18--33. Springer.

\bibitem[{Hunter and Thimm(2016)}]{HunterThimm2016ijar}
Hunter, A.; and Thimm, M. 2016.
\newblock Optimization of Dialectical Outcomes in Dialogical Argumentation.
\newblock \emph{International Journal of Approximate Reasoning}, 78: 73--102.

\bibitem[{Jannach et~al.(2021)Jannach, Manzoor, Cai, and Chen}]{JannachMCC21}
Jannach, D.; Manzoor, A.; Cai, W.; and Chen, L. 2021.
\newblock A Survey on Conversational Recommender Systems.
\newblock \emph{{ACM} Comput. Surv.}, 54(5): 105:1--105:36.

\bibitem[{Katsumi et~al.(2018)Katsumi, Hiraoka, Yoshino, Yamamoto, Motoura,
  Sadamasa, and Nakamura}]{Katsumi2018}
Katsumi, H.; Hiraoka, T.; Yoshino, K.; Yamamoto, K.; Motoura, S.; Sadamasa, K.;
  and Nakamura, S. 2018.
\newblock Optimization of Information-Seeking Dialogue Strategy for
  Argumentation-Based Dialogue System.
\newblock In \emph{Proceedings of {DEEP-DIAL@AAAI'19}}, volume abs/1811.10728.
  {ArXiv}.

\bibitem[{Lenert et~al.(1997)Lenert, Morss, Goldstein, Bergen, Faustman, and
  Garber}]{lenert1997measurement}
Lenert, L.; Morss, S.; Goldstein, M.~K.; Bergen, M.; Faustman, W.; and Garber,
  A.~M. 1997.
\newblock Measurement of the validity of utility elicitations performed by
  computerized interview.
\newblock \emph{Medical Care}, 35(9): 915--920.

\bibitem[{Lenert, Sherbourne, and Reyna(2001)}]{lenert2001utility}
Lenert, L.~A.; Sherbourne, C.~D.; and Reyna, V. 2001.
\newblock Utility elicitation using single-item questions compared with a
  computerized interview.
\newblock \emph{Medical Decision Making}, 21(2): 97--104.

\bibitem[{Monteserin and Amandi(2013)}]{Monteserin2012}
Monteserin, A.; and Amandi, A. 2013.
\newblock A reinforcement learning approach to improve the argument selection
  effectiveness in argumentation-based negotiation.
\newblock \emph{Expert Systems with Applications}, 40: 2182--2188.

\bibitem[{Osborne and Rubinstein(1994)}]{Osborne1994}
Osborne, M.; and Rubinstein, A. 1994.
\newblock \emph{A Course in Game Theory}.
\newblock {MIT} Press.

\bibitem[{Peterson(2009)}]{Peterson1994}
Peterson, M. 2009.
\newblock \emph{An Introduction to Decision Theory}.
\newblock Cambridge University Press.

\bibitem[{Rach, Minker, and Ultes(2018)}]{Rach2018}
Rach, N.; Minker, W.; and Ultes, S. 2018.
\newblock Markov Games for Persuasive Dialogue.
\newblock In \emph{{COMMA}}, volume 305 of \emph{Frontiers in Artificial
  Intelligence and Applications}, 213--220. {IOS} Press.

\bibitem[{Rahwan and Larson(2008)}]{RahwanLarson08}
Rahwan, I.; and Larson, K. 2008.
\newblock Pareto Optimality in Abstract Argumentation.
\newblock In \emph{{AAAI}}, 150--155. {AAAI} Press.

\bibitem[{Rahwan, Larson, and Tohm{\'{e}}(2009)}]{RL09}
Rahwan, I.; Larson, K.; and Tohm{\'{e}}, F.~A. 2009.
\newblock A Characterisation of Strategy-Proofness for Grounded Argumentation
  Semantics.
\newblock In \emph{{IJCAI}}, 251--256.

\bibitem[{Ricci, Rokach, and Shapira(2015)}]{2015rsh}
Ricci, F.; Rokach, L.; and Shapira, B., eds. 2015.
\newblock \emph{Recommender Systems Handbook}.
\newblock Springer.
\newblock ISBN 978-1-4899-7636-9.

\bibitem[{Rienstra, Thimm, and Oren(2013)}]{RienstraThimmOren13}
Rienstra, T.; Thimm, M.; and Oren, N. 2013.
\newblock Opponent Models with Uncertainty for Strategic Argumentation.
\newblock In \emph{{IJCAI}}, 332--338. {IJCAI/AAAI}.

\bibitem[{Riveret et~al.(2019)Riveret, Gao, Governatori, Rotolo, Pitt, and
  Sartor}]{Riveret2019}
Riveret, R.; Gao, Y.; Governatori, G.; Rotolo, A.; Pitt, J.; and Sartor, G.
  2019.
\newblock A probabilistic argumentation framework for reinforcement learning
  agents - Towards a mentalistic approach to agent profiles.
\newblock \emph{Autonomous Agents and Multi-Agent Systems}, 33(1-2): 216--274.

\bibitem[{Rosenfeld and Kraus(2016{\natexlab{a}})}]{Rosenfeld2016}
Rosenfeld, A.; and Kraus, S. 2016{\natexlab{a}}.
\newblock Providing Arguments in Discussions on the Basis of the Prediction of
  Human Argumentative Behavior.
\newblock \emph{{ACM} Transactions on Interactive Intelligent Systems}, 6(4):
  30:1--30:33.

\bibitem[{Rosenfeld and Kraus(2016{\natexlab{b}})}]{Rosenfeld2016ecai}
Rosenfeld, A.; and Kraus, S. 2016{\natexlab{b}}.
\newblock Strategical Argumentative Agent for Human Persuasion.
\newblock In \emph{{ECAI}}, volume 285 of \emph{Frontiers in Artificial
  Intelligence and Applications}, 320--328. {IOS} Press.

\bibitem[{Stankevitz et~al.(2017)Stankevitz, Dement, Schoenfisch, Joyner,
  Clancy, Stroo, and {\O}stbye}]{stankevitz2017perceived}
Stankevitz, K.; Dement, J.; Schoenfisch, A.; Joyner, J.; Clancy, S.~M.; Stroo,
  M.; and {\O}stbye, T. 2017.
\newblock Perceived barriers to healthy eating and physical activity among
  participants in a workplace obesity intervention.
\newblock \emph{Journal of Occupational and Environmental Medicine}, 59(8):
  746--751.

\bibitem[{Turk and Borkowski(2005)}]{turk2005review}
Turk, P.; and Borkowski, J.~J. 2005.
\newblock A review of adaptive cluster sampling: 1990--2003.
\newblock \emph{Environmental and Ecological Statistics}, 12(1): 55--94.

\end{thebibliography}
\end{document}